\documentclass[twocolumn,11pt]{article}
\usepackage{times}
\usepackage{graphicx}
\usepackage{algorithm}
\usepackage{algpseudocode}
\usepackage{array}
\usepackage{booktabs}
\usepackage{url}
\usepackage{hyperref}
\usepackage{amssymb}
\usepackage{amsmath}

\begin{document}
\global\def\refname{{\normalsize \it References:}}
\baselineskip 12.5pt
%
%
% TITLE, AUTHOR, ABSTRACT, KEYWORDS
%
\title{\LARGE \bf Real-time Framework for Interoperable Semantic-driven Internet-of-Things in Smart Agriculture}

\date{}

\author{\hspace*{-10pt}
\begin{minipage}[t]{2.7in} \normalsize \baselineskip 12.5pt
\centerline{MOHAMED EL-DOSUKY}
\centerline{Computer Science Department, Arab East Colleges, Saudi Arabia}
\centerline{Computer Science Department, Faculty of Computers and Information, Mansoura University, Egypt}
\centerline{maldosuky@arabeast.edu.sa}
\end{minipage} \kern 0in
%\begin{minipage}[t]{2.7in} \normalsize \baselineskip 12.5pt
%\centerline{SECOND AUTHOR}
%\centerline{Name of the University}
%\centerline{Institute of Mechanical Engineering}
%\centerline{47 West Lincoln Avenue, 87 115 City}
%\centerline{COUNTRY}
%\centerline{second.author@math.univ.ab}
%\end{minipage}
%
% If you are three authors then you can use three mini--pages
% instead of two. Their horizontal size must be less than 2.7in
% indicated above. It can be e.g. 2.3in. However, you must pay
% attention that you do not exceed the total width of the text.
%
\\ \\ \hspace*{-10pt}
\begin{minipage}[b]{6.9in} \normalsize
\baselineskip 12.5pt {\it Abstract:}
% The text of the abstract follows.
The Internet of Things (IoT) has revolutionized various applications including agriculture, but it still faces challenges in data collection and understanding. This paper proposes a real-time framework with three additional semantic layers to help IoT devices and sensors comprehend data meaning and source. The framework consists of six layers: perception, semantic annotation, interoperability, transportation, semantic reasoning, and application, suitable for dynamic environments. Sensors collect data in the form of voltage, which is then processed by microprocessors or microcontrollers in the semantic annotation and preprocessing layer. Metadata is added to the raw data, including the purpose, ID number, and application. Two semantic algorithms are proposed in the semantic interoperability and ontologies layer: the interoperability semantic algorithm for standardizing file types and the synonym identification algorithm for identifying synonyms. In the transportation layer, raw data and metadata are sent to other IoT devices or cloud computing platforms using techniques like WiFi, Zigbee networks, Bluetooth, and mobile communication networks. A semantic reasoning layer is proposed to infer new knowledge from the existing data, using fuzzy logic, Dempster-Shafer theory, and Bayesian networks.  A Graphical User Interface (GUI) is proposed in the application layer to help users communicate with and monitor IoT sensors, devices, and new knowledge inferred. This framework provides a robust solution for managing IoT data, ensuring semantic completeness, and enabling real-time knowledge inference. The integration of uncertainty reasoning methods and semantic interoperability techniques makes this framework a valuable tool for advancing IoT applications in general and in agriculture in particular.
\\ [4mm] {\it Key--Words:}
% The key-words follow.
Interoperability, Semantic, Internet-of-Things, Agriculture
\end{minipage}
\vspace{-10pt}}

\maketitle

\thispagestyle{empty} \pagestyle{empty}
% numbers of pages are supplemented by the editor
%
% THE BEGINNING OF THE TEXT
%
\section{Introduction}
\label{S1} \vspace{-4pt}

The Internet of Things (IoT) is an innovative technology significantly changing how we interact with the physical world. It's no longer just about linking our phones and laptops to the internet; it's about creating a network of interconnected gadgets that can gather, distribute, and analyze data, resulting in a smarter, more automated society. Its fundamental goal is to enhance convenience and efficiency breakthrough that provides efficient use in Information and Communication Technologies (ICT) \cite{nolin2016internet}.

Smart agriculture is one of the domains that benefit from IoT \cite{sinha2022recent}. IoT devices can help farmers optimize their resources, improve crop yields, and reduce waste through precision farming techniques \cite{sanjeevi2020precision}.

Despite the efficiency of IoT devices, they unfortunately lack the ability to understand the meaning of the data collected by their sensors. Therefore, it is important to integrate IoT with semantic web techniques.   The Semantic Web community is important for ensuring data interoperability and integrating findings. Semantic analytics is a new endeavor that uses Linked Open Data reasoning to extract meaningful information from IoT data. It aims to create new knowledge through interoperable data interpretation \cite{hitzler2021review}.

While massive amounts of data hold the key to uncovering hidden patterns (through data mining) that can improve our lives, building truly intelligent machines that can perceive the world requires integrating data from all our interconnected devices using existing technology \cite{chander2022artificial}. To address this challenge, the World Wide Web Consortium (W3C) established a dedicated group (Web of Things Community Group) that developed the oneM2M standards. These standards aim to create a common language for devices to share data seamlessly, leveraging familiar web technologies like RESTful, HTTP, and RDF, ultimately paving the way for integrating IoT data into AI for a future filled with intelligent machines \cite{nivzetic2020internet}. Semantic analytics integrates several technologies and analytic tools, including logic-based reasoning, machine learning, and Linked Open Data, to transform data into actionable information. It blends semantic web technologies with reasoning methodologies \cite{gil2016internet}.

As the number of IoT devices increases every year, each device collects vast amounts of data every minute, making it critical to understand this data beyond static methods \cite{statista}. Cisco reports that 4.8 zettabytes of data are transferred annually without any information about the data's description \cite{cisco2020cisco}.
%Additionally, Gartner indicates that 70\% of IoT projects face interoperability issues due to the use of specific data formats and communication protocols by each device [10].
McKinsey's research highlights that the productivity and performance of IoT systems can be increased by 30\% by integrating semantic methods into IoT systems \cite{bughin2017case}. Therefore, it is important to integrate IoT with semantic web techniques.

The objective of this paper is to propose a new framework that adds semantic techniques to IoT platforms in general and in agriculture as a case study. This framework aims to identify the source of the data, the location of the sensor that sent the data, understand the meaning of each word, and suggest synonyms to explain its meaning. This framework also provides information and meaning for each piece of data collected by the sensors.

The rest of the paper is organized as follows: In Section \ref{sec:Background}, a background on IoT and its application in agriculture are presented, and the different types of semantic methods that can be added to the IoT environment are discussed. In Section \ref{sec:Related}, some recent works are introduced. The proposed framework is presented and discussed in Section \ref{sec:Methodology}. Section \ref{sec:Agriculture} provides a case study in applying the framework in agriculture. Finally, in Section \ref{sec:Conclusion}, the conclusion of this paper is presented with some directions for future research.

\section{Background}
\label{sec:Background} \vspace{-4pt}

IoT is considered the third wave of internet technology \cite{tripathy2017internet}. The first wave involved connecting two computers together, and the second wave was the World Wide Web (WWW) . IoT is a technology that enables any object to transmit and receive data over the internet \cite{sadeghi2023internet}. Nowadays, IoT devices can transmit and receive data to and from other IoT devices or upload and download it to cloud computing \cite{ahlawat2023towards}. To enable IoT devices to interact with their environment independently, they consist of three main components: sensors, actuators, and a brain device \cite{laghari2021review}.

A sensor is a device that can retrieve and record environmental measurements such as temperature, humidity, and distance. Actuators enable the IoT device to perform specific tasks in the environment, such as using a motor to open or close a door. The brain device, which can be a microcontroller like the Arduino Uno or a microprocessor like the Raspberry Pi, takes the environmental measurements from the sensors, processes them into meaningful data, and then sends the necessary actions to the actuators to perform specific tasks. As a result, IoT devices can send and receive data over the internet without human intervention \cite{zikria2021next}.

\subsection{IoT application in Agriculture} \vspace{-4pt}
Regarding agriculture, the integration of the Internet of Things (IoT) into agriculture has revolutionized farming practices by providing real-time data and insights that empower farmers to make informed decisions \cite{castrignano2020agricultural}. One of the key advancements is the ability to monitor crucial environmental parameters such as temperature, humidity, and soil moisture levels \cite{sinha2022recent}. By using a network of IoT sensors deployed throughout their fields, farmers can continuously collect and analyze this data, allowing them to understand the specific conditions affecting their crops at any given moment \cite{ayaz2019internet}.

IoT technology allows farmers to optimize their operations by monitoring soil moisture levels, determining irrigation timing, and analyzing temperature and humidity data \cite{obaideen2022overview}. This helps conserve water and promotes healthier plant growth.
IoT technology can also predict weather patterns, adjusting farming practices accordingly \cite{fuentes2024transformative}. It can recommend effective irrigation methods based on soil type and crop requirements, enhancing crop yields while minimizing waste \cite{shah2019soil}. IoT solutions also facilitate better pest and disease management by integrating environmental data with pest life cycles, reducing reliance on chemical pesticides and promoting sustainable farming practices \cite{angon2023integrated}.
\subsection{Semantic IoT} \vspace{-4pt}
Implementing a complete IoT system must consist of three components: things-oriented, internet-oriented, and semantic-oriented. The things-oriented component is related to sensors and how they collect data from the environment. The internet-oriented component relates to how these sensors communicate over the internet and share data. Sensors share data without understanding its meaning, so the semantic-oriented component focuses on how these sensors understand the data and provide synonyms for any concept without human intervention in any application that IoT introduces. To enable IoT systems to understand the meaning and synonyms of the concepts in the shared data, the IoT field must be combined with knowledge engineering and machine learning \cite{rhayem2020semantic}.

According to Shewale's report, there are 17.08 billion IoT devices in use worldwide, and it is predicted that by the end of 2030, there will be 29.42 billion IoT devices around the world \cite{demandsage}. These devices lack many semantic services such as interoperability, automatic data reception and transmission, synonym identification for concepts, and intelligent communication. These services help IoT users easily discover, understand, store, and track IoT data. To add these services to IoT devices, new concepts such as semantic interoperability, ontologies, linked open data, semantic reasoning and annotation must be developed and designed for IoT systems. When semantic techniques are implemented in IoT devices, the IoT system is called the Internet of Everything (IoE), and the semantic techniques are referred to as the Semantic Web of Things \cite{farias2021internet, rhayem2020semantic}.
\subsubsection{Semantic interoperability} \vspace{-4pt}
Interoperability in IoT systems means the ability to exchange data between two computers or devices. IoT devices often lack understanding of data representation, as they primarily understand XML, JSON, and CSV formats, while other types are considered incompatible. They also struggle with understanding the meaning of the data and the methods of data transfer between devices. Semantic techniques can be added to interoperability to enable devices or computers not only to exchange data but also to understand the importance and meaning of the data being transferred. This approach standardizes the way of transferring data, allowing devices to send, receive, and comprehend data in any format and representation. Semantic interoperability also provides common and accepted vocabularies to share the meaning of the data between devices, including techniques that understand synonyms of words \cite{novo2020semantic}.
\subsubsection{Ontologies} \vspace{-4pt}
Ontologies refer to the ability to describe knowledge using concepts or classes for each word and the relationships between them. Because of the dynamic environment of IoT, the term ontology is used to represent and share knowledge among different applications. The use of ontologies in the IoT field aims to describe knowledge that comes from sensors as a set of concepts or classes, defining the properties of each class and finding the relationships between classes. To achieve this, the Resource Description Framework (RDF) is used to describe ontologies using directed graph techniques \cite{qaswar2022applications}.
\subsubsection{Linked open data} \vspace{-4pt}
The Semantic Web is an advanced version of the World Wide Web (WWW) that aims to provide human- and machine-readable content across the internet. This helps both machines and individuals access any data on the internet by building a framework that allows data to be reused among applications. The concept of the Semantic Web originated from linked data, which aims to create links connecting data across the internet. These links are human- and machine-readable, meaning they contain necessary information. Uniform Resource Identifiers (URIs) are used as identifiers to provide information about the content of any website. Linked open data follows the same concept as linked data but focuses on providing information about links that include free data \cite{pauwels2021knowledge}.
\subsubsection{Semantic reasoning} \vspace{-4pt}
Semantic reasoning is a type of reasoning applied in knowledge-based systems to infer new knowledge from existing knowledge. The simplest form of semantic reasoning uses IF-THEN rules, such as SPARQL Inferencing Notation and Semantic Web Rule Language. In the IoT environment, the IF-THEN rule alone is not sufficient to understand and infer new knowledge from existing data because IoT systems collect data from many sensors distributed across various areas. Therefore, IoT requires distributed reasoning, which aims to gather information from multiple sensors, store it on a server, and then apply IF-THEN rules or other semantic reasoning techniques to infer new knowledge. Distributed reasoning enhances the efficiency of IoT systems by minimizing latency in data collection and processing, thereby improving the system's ability to infer new knowledge quickly and accurately \cite{zgheib2023scalable}.
\subsubsection{Semantic annotation} \vspace{-4pt}
Semantic annotations involve adding metadata to IoT devices by providing descriptions for each device, sensor, and any related phenomena. These annotations help in identifying each sensor in the IoT environment, distinguishing them, and identifying the data sources. Semantic annotations facilitate easier interoperability between devices and sensors. In IoT, semantic annotations can appear as mapped representations of sensor locations along with information about each sensor and device. D2RQ and R2RML are languages that can be used to create these maps of sensor information \cite{chikkamannur2020semantic}.

\section{Related Work}
\label{sec:Related} \vspace{-4pt}
One of the most significant limitations of IoT devices is their inability to understand the data collected by sensors. As a result, many recent works focus on the importance of adding semantics to IoT devices.
%For example, Guerroudj and Siam [33] present the potential of IoT technology and highlight the importance of implementing semantic web technology in IoT systems. They also show how semantic IoT can be applied across various IoT applications such as healthcare systems, smart cities, and smart homes. Additionally, they discuss several semantic techniques that can be appropriate for application in IoT systems.

Palo \cite{palo2021semantic} highlights the importance of IoT and the services it introduces. He also explains how data can be transmitted and shared between devices without understanding its meaning or importance. To address this, he introduces methods for enabling IoT devices to understand the data they transmit. The author focuses on semantic interoperability, showcasing the advantages and disadvantages of applying semantic interoperability in IoT systems. He also discusses how to secure IoT devices after implementing semantic techniques.
Ganzha et al. \cite{ganzha2017semantic} explains how ontologies and semantic data processing might improve interoperability in the IoT world.
Pliatsios et al. \cite{pliatsios2023systematic} provide a literature review on how semantic techniques can be applied to achieve semantic interoperability in smart cities, an application of IoT. They address the challenges of applying semantic interoperability techniques in green and sustainable smart cities. Additionally, they discuss recent works that demonstrate how various semantic techniques, such as linked open data, automatic reasoning, and knowledge graphs, can be applied in smart cities to help sensors and IoT devices understand the data they share.

Nagasundaram et al. \cite{nagasundaram2024proposed} described a fog-based conceptual paradigm to enable dynamic semantic interoperability in IoT. The model includes a single-tier fog layer, which provides the necessary computing power to accomplish this aim. They offer a complete literature analysis on semantic interoperability, focusing on latency, bandwidth, overall cost, and energy use.
Many recent works suggest new architectures and frameworks to add semantic techniques to the standard IoT layers. For example, Yao et al. \cite{yao2024semantic} proposed a brief framework of semantic processing for interoperability in the Industrial Internet of Things, featuring task orientation and collaborative processing. They achieve a more efficient way of processing and exchanging information than conventional methods, which is critical for handling the demands of future interconnected industrial networks.

Mondrag\'{o}n et al. \cite{mondragon2021experimental} proposed a federated fog computing relies heavily on semantic interoperability to enable smooth integration and interaction across diverse IoT devices and fog nodes.
Mante \cite{mante2023iot} proposed a new IoT architecture that incorporates semantic techniques to enable devices and sensors to understand the data they share. This architecture can be applied in smart cities and consists of four layers. The first layer focuses on monitoring and collecting data from the system using the oneM2M concept. The second layer suggests storing and sending data between devices and sensors using interoperability techniques. In the third layer, Mante proposes monitoring the energy and electricity required for vehicle charging. The fourth layer uses open API and IUDX standard data schema to exchange data between systems. Finally, the authors recommend using semantic techniques to add meaning and descriptions to the data collected by the sensors.

Souza et al. \cite{de2022semantic} proposed a method relies heavily on semantic technologies like ontologies, the Resource Description Framework (RDF), and reasoning procedures to provide a common understanding of data and promote data exchange, discovery, and integration across domains.
Interestingly, the interoperability solutions mentioned above, either do not use semantic methods at all, or use them rather sparingly. Ontologies are used in architectures at the IoT stack's lowest level. Herzog and Buchmann's \cite{herzog2012a3me} A3ME middleware represents devices in heterogeneous sensor/actuator networks as agents. The A3ME architecture allows for device detection, semantic description sharing, and basic interactions across devices. Kiljander et al. \cite{kiljander2014semantic} propose an interoperability architecture at the sensor/perception layer. Edgar et al. \cite{huaranga2024cloud} proposed a comparative analysis of computational resources in real-time IoT applications based on semantic interoperability. They discuss challenges in attaining semantic interoperability across cloud, fog, and federated-fog computing systems.

To increase the efficiency of semantic IoT techniques, many recent works use machine learning algorithms to add semantics to IoT devices. For example. Rahman et al. \cite{rahman2023dyno} used machine learning techniques to develop and design a semantic method for IoT systems. This semantic method can be used in IoT environments with dynamic behavior. To add semantics to IoT systems, the authors rely on ontology methodology techniques. Their method can be applied in various IoT applications such as smart cities, smart homes, and healthcare systems. The authors conducted experiments that showed their system outperformed other existing ontology techniques, achieving accuracy 17\% higher than existing technology. The experimental results also demonstrated that their system decreased response time to queries by 35\% compared to existing ontology systems.

Di Martino and Esposito \cite{dimartino2021semantic} proposed a prototype tool to add semantic interoperability in the IoT environment. This tool, called REST API, relies on semantic graph techniques to integrate semantics into IoT. It involves analyzing data and then providing a semantic explanation of this data. The tool operates through three main steps to provide the meaning of the data: critical analysis, manual annotation, and production phase. In the analysis phase, the authors used automatic and manual tools to analyze the data collected by sensors. Each word is categorized, and the relationship between words is identified using manual annotation tools in the annotation phase. Finally, the meaning of the data is provided in the production phase. The authors also highlight the importance of adding semantics to the IoT environment and discuss different semantic methodologies that can be integrated into IoT systems.

Lynda et al. \cite{lynda2023towards} proposed a machine learning system aimed at identifying the meaning of data collected by IoT systems and determining whether this data is related to agriculture. The system operates through three main steps: perception, semantics, and classification. In the perception phase, many sensors collect various data. This data is then analyzed to understand its meaning using ontology methods. Machine learning classifiers such as support vector machines, k-nearest neighbors, decision trees, and naive Bayes are then used to classify whether the data is related to agriculture or not. The experimental results show that the support vector machine outperformed other classifiers, achieving 99\% accuracy.

Guleria and Sood \cite{guleria2021semantic} discuss the different methodologies that can add meaning to the data collected by IoT sensors. They also highlight the role of the semantic web in adding meaning to data on the web and show that the IoT framework has three layers to incorporate semantic interoperability into IoT applications. The authors also proposed a new machine learning-based text analytics model that can be used in healthcare systems to perform semantic classification of medical data.

While several prior works have proposed semantic-aware IoT architectures or employed ontologies and reasoning to improve interoperability (e.g., \cite{mante2023iot}, Souza et al. \cite{de2022semantic}, Rahman et al.\cite{rahman2023dyno}), the proposed framework differs in three concrete ways that together create a distinct, real-time solution for smart agriculture. First, rather than treating semantics as an orthogonal add-on, the framework embeds three dedicated semantic layers (semantic annotation, semantic interoperability \& ontologies, and semantic reasoning) into a single operational pipeline designed for low-latency environments; this vertical integration ensures that annotation, mapping and inference operate as coordinated, streaming steps rather than isolated batch tasks. Second, our semantic annotation is intentionally non-ML and deterministic: it performs format-agnostic, rule-based enrichment at line-rate so devices and gateways with limited compute can immediately produce semantically tagged streams without the training/maintenance overhead of ML models. Third, the interoperability and synonym-identification algorithms introduced for Layer 3 are lightweight, uncertainty-aware, and format-agnostic — they explicitly reconcile heterogeneous payloads (JSON, CSV, telemetry frames, simple CSV/TSV, etc.) on the fly and propagate confidence scores into the reasoning layer to support uncertainty-based decisions. In contrast, the cited architectures either focus on ontology design and offline reasoning, rely heavily on ML-based semantic extraction, or assume homogeneous message formats; they therefore cannot guarantee the same combination of real-time throughput, minimal device-side requirements, and uncertainty-aware inference that our design targets. Taken together, these design choices enable practical deployment across constrained agricultural devices and gateways while preserving semantic fidelity for higher-level analytics and control.

\section{Methodology}
\label{sec:Methodology} \vspace{-4pt}

\begin{figure}[t]
\centering
\includegraphics[width=0.5\textwidth]{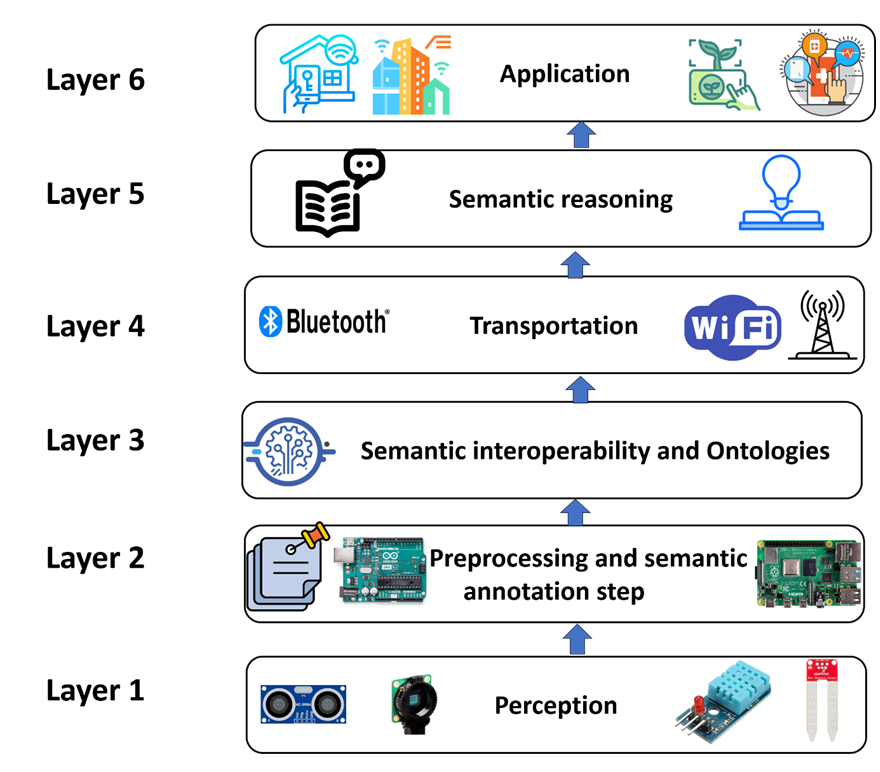}
\caption{Proposed IoT framework}\label{fig:framework}
\end{figure}

In this paper, a new IoT framework is proposed to add semantic representation to the data collected through sensors. This framework can be applied to any IoT application. It can easily be used in dynamic environments as it employs real-time techniques to add semantics to the IoT-collected data. The proposed framework is shown in Figure \ref{fig:framework}. It illustrates that the framework comprises six layers: perception, semantic annotation and preprocessing, semantic interoperability and ontologies, transportation, semantic reasoning, and application. To enable IoT sensors and devices to understand the data they share, three layers—semantic annotation, semantic interoperability and ontologies, and semantic reasoning—are added in the proposed framework to incorporate the three types of semantics in the IoT.

\subsection{Perception layer}
The perception layer contains IoT sensors and is responsible for collecting environmental data. It includes hundreds of sensors such as temperature, humidity, ultrasonic, and heartbeat sensors. There are two types of sensors: passive and active, which collect data from the environment. Passive sensors do not emit any rays into the environment; they collect data by detecting signals emitted from objects in the environment, such as thermostats and LM35 sensors. Active sensors, on the other hand, emit signals into the environment, and these signals are reflected back to the sensor with the measured data. For example, an ultrasonic sensor emits a signal at a specific speed, and when this signal returns to the sensor, it can easily compute the distance between objects. Both types of sensors collect data in the form of electrical signals (voltage), which is not readable by humans. So, this layer called physical layer. The main challenges of this layer are how to secure the sensors and the data they collect. There are many types of attacks that this layer faces, such as:
\begin{itemize}
    \item \textbf{Eavesdropping:} In this type of attack, the attackers create a private network with the sensors, making it easy to steal the data that the sensors collect.
    \item \textbf{Node Capture:} The attackers gain full access and control of key elements in the network of the sensors, such as gateways. They can then easily obtain the data sent or received over the network from the memory.
    \item \textbf{Fake Node:} In this type of attack, the attackers create a new node in the wireless sensor network that can communicate with other devices and sensors to send and receive data from them, and then steal it.
    \item \textbf{Timing Attack:} In this kind of attack, the hackers search for weak devices in the wireless sensor network with low computing capability. They identify vulnerable points to attack and steal information from these devices. The attackers measure the time it will take to attack these devices.
\end{itemize}

To address these challenges, it is crucial to implement robust security measures in the perception layer, such as encryption, authentication, and intrusion detection systems, to protect the integrity and confidentiality of the data collected by IoT sensors. These measures help ensure that the data remains secure and that unauthorized access is prevented, thereby maintaining the overall reliability and trustworthiness of the IoT system.

\subsection{semantic annotation and preprocessing layer} \vspace{-4pt}
After the sensors collect data, they send it to the semantic annotation and preprocessing layer. Sensors measure and collect data in the form of voltage, so in the preprocessing layer, the data is converted into meaningful information. A microcontroller like the Arduino Uno or a microprocessor like the Raspberry Pi can be used to preprocess the voltage data into meaningful data. These devices can also be used to perform a new real-time semantic annotation technique.
This layer aims to help the IoT system identify the source of the data and the application for which these sensors collected the data. It also aims to provide a descriptive explanation for the data collected by these sensors. To perform semantic annotation in this layer, we suggest a new real-time technique that assists sensors in adding metadata after they collect the data. This metadata includes the sensor ID and a full description of the data collected by the sensor. Each sensor is assumed to automatically embed this metadata into the data before sending it to the microcontroller device. Additionally, GPS is suggested for identifying the location of each sensor and including this information in the collected data.

\begin{figure}[t]
\centering
\includegraphics[width=0.5\textwidth]{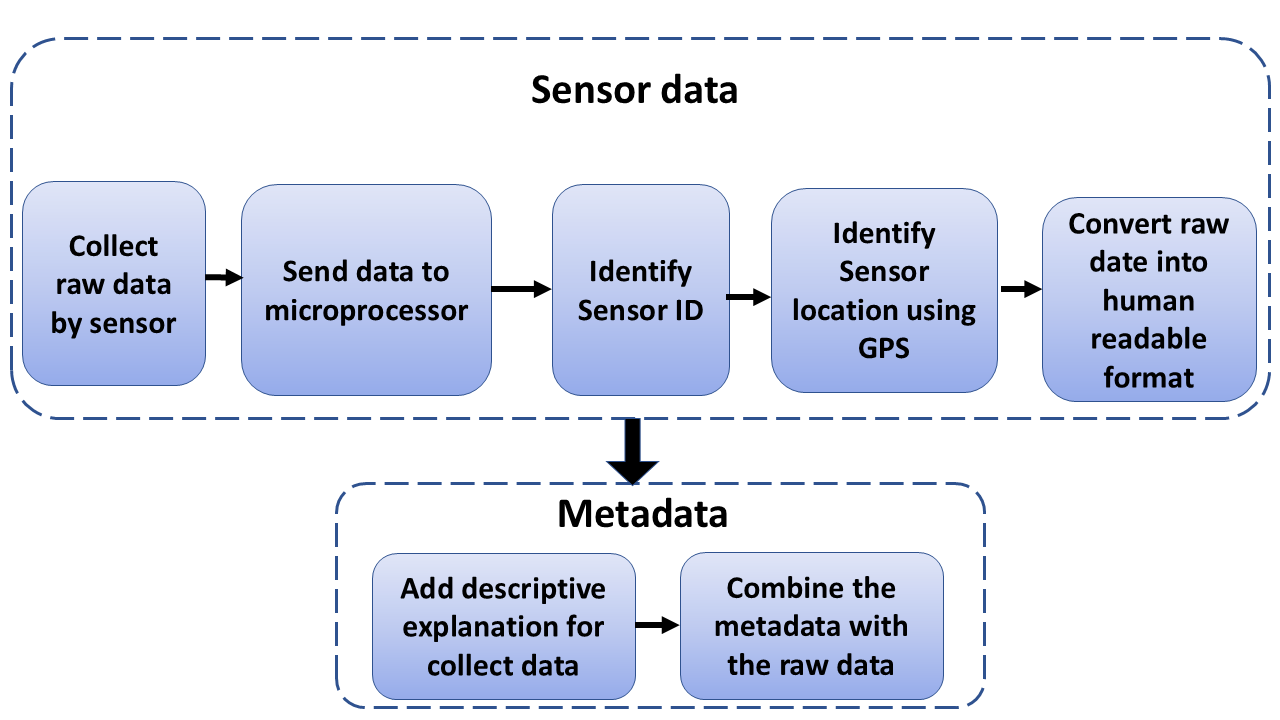}
\caption{Semantic annotation steps}\label{fig:meta}
\end{figure}

Figure \ref{fig:meta} shows the steps that we propose to add real-time annotation semantic techniques without the need for machine learning and artificial intelligence techniques, which can consume time. The real-time semantic technique consists of two main steps: sensor data and metadata. In the sensor data step, the sensor sends the raw data it collected to the microprocessor or microcontroller. The microprocessor or microcontroller then identifies the sensor using an ID that consists of three main components. The first component is the job that this sensor performs. The second part contains the ID of this sensor. The third part indicates the application that this sensor serves. For example, if a sensor ID is TEMP102SC, this means that the sensor measures the temperature of the environment, has ID 102, and serves the smart city application. Additionally, in this step, the location of the sensor is estimated using the Global Positioning System (GPS). The final step in the sensor data step is converting the actual data collected from the environment into a human-readable form.
In the metadata step, a descriptive explanation of the data measured by the sensor is added. Thus, the raw data collected by the sensor, along with the metadata (ID, location of the sensor, and description of the data), are combined to form new data that includes both the data and its explanation.

\subsection{Semantic interoperability and ontologies layer} 
\begin{figure*}[t]
\centering
\includegraphics[width=0.75\textwidth]{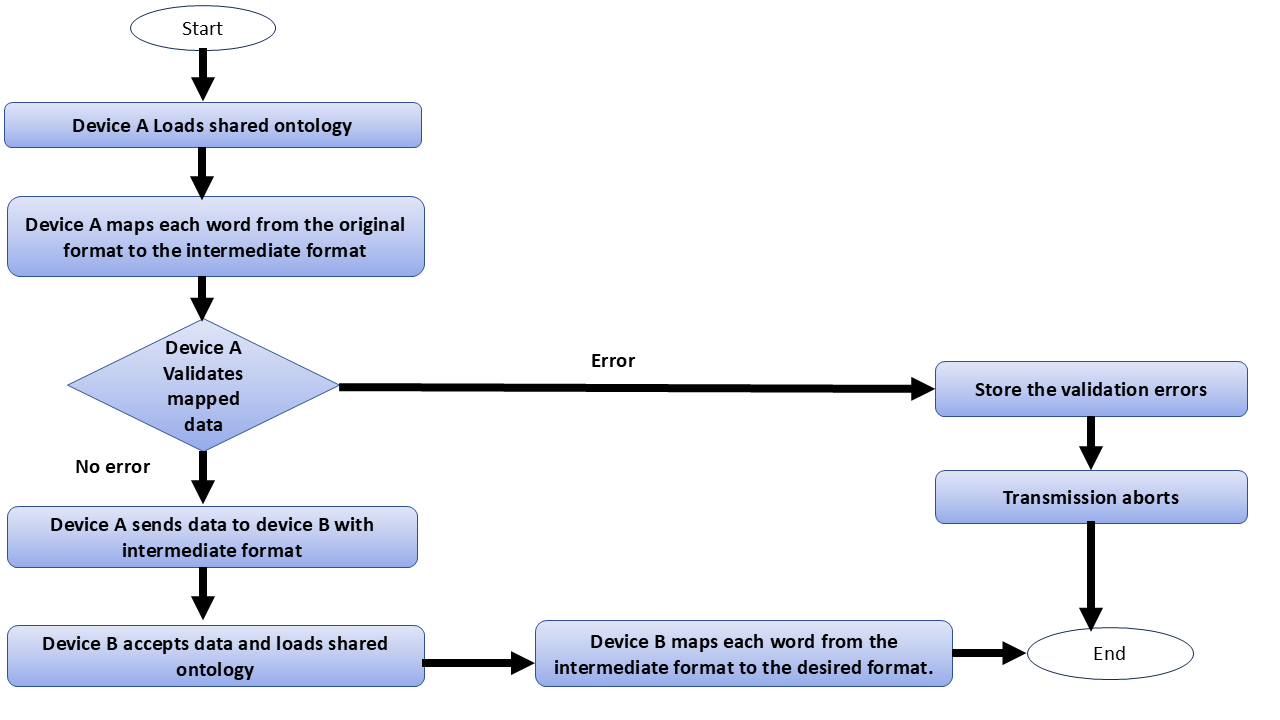}
\caption{Semantic interoperability flowchart}\label{fig:devices}
\end{figure*}

\vspace{-4pt}
The data from the semantic annotation and preprocessing layer provides a full understanding of the data collected by sensors, but it lacks a standardized format. Additionally, this data may contain synonymous words that are unfamiliar to many people. Therefore, semantic interoperability techniques are incorporated into this layer to allow devices to share collected data in any format and using any protocols for transportation. To achieve this, a new real-time semantic interoperability technique is proposed to enable IoT devices to share data without a specific format and to suggest synonyms for words that may be difficult for some people to understand.
So, in this layer, two main algorithms are proposed: the semantic interoperability algorithm and the synonymous identification algorithm. These algorithms help IoT devices share data without a specific format and facilitate understanding of unfamiliar words by finding synonyms.
The semantic interoperability algorithm addresses the problem of IoT devices requiring specific data formats such as XML, JSON, and CSV. The flowchart for the semantic interoperability algorithm is shown in Figure \ref{fig:devices}. It illustrates that a shared ontology is used to help any IoT device transform any data format to another format. It also shows that the device sending the data must first transform the data format to an intermediate format that any IoT device can accept by mapping each word in the original file to the intermediate format. The sending device must validate the data to ensure it is transformed without errors. If the sending device maps the data without errors, it sends the data to the receiving device, which uses the shared ontology to transform the intermediate data format to its specific desired format. If the sending device discovers an error during validation, it records the error and stops transmitting the data.

A synonymous identification algorithm is also proposed in this layer to find the synonyms of each word, as many words have numerous synonyms that individuals might not be aware of. Algorithm 1 outlines the proposed synonymous identification algorithm. It shows that the algorithm follows three main steps: removing stop words, finding synonyms, and converting synonyms to their base form. In the remove stop words step of the algorithm, stop words are removed because these words do not have synonyms. Additionally, the non-stop words are converted to their root forms to facilitate the identification of synonyms. In the find synonyms step, the WordNet database is used to retrieve a list of synonyms for each word. Finally, in the convert synonyms to base form step, each synonym is converted to its lemma (base) form.

\begin{algorithm}
\caption{Synonymous Identification Algorithm} \label{algo:Syn}
\begin{algorithmic}[1]
\State \textbf{Input:} Original data \( O \)
\State \textbf{Output:} Synonymous word matrix \( SW \)
\State Break the original data into words: \( Words = [``word1'', ``word2'', ``word3'', \dots] \)
\State Final\_keywords = [ ]
\State \( SW = [ ] [ ] \)
\For{$i = 1$ to $Words.length$}
    \If{$Word[i] \neq$ stop word}
        \State Convert $Word[i]$ to its stemmed form
        \State Final\_keywords.append($Word[i]$)
    \EndIf
\EndFor
\For{$i = 1$ to $Words.length$}
    \State Use WordNet database to find the synonym set for each word and call it synsets
    \For{$j = 1$ to $synsets.length$}
        \State Convert synonymous synsets[$j$] to lemma form $lemma[j]$
        \State $SW[i][j] = lemma[j]$
    \EndFor
\EndFor
\State \Return $SW$
\end{algorithmic}
\end{algorithm}

\subsection{Transportation layer} \vspace{-4pt}
After preprocessing, the final data must be transferred to another IoT system, the internet, or a database for storage and processing. The collected IoT data can also be sent to and stored in a cloud data center in this layer. This layer relies on sending data using IP addresses and port numbers. Various methods can be used for transmitting and receiving data, including Wi-Fi, radio-frequency identification (RFID), Bluetooth, 3G, 4G, and Zigbee networks. These methods use different data transfer protocols to transmit data from one IoT device to another or to a data center over the internet.

%Table 1 shows the most important data transfer protocols used in IoT systems. These protocols vary in terms of data transmission capacity, transmission speed, data security, and power consumption during transmission.
Let us compare important IoT communication protocols. Message Queue Telemetry Transport (MQTT) is a widely used, open-source protocol known for its lightweight nature, utilizing a publish-subscribe mechanism to ensure efficient data transmission even in low-connectivity environments. Modbus is primarily used for connecting supervisory systems to remote terminals, allowing data acquisition and control, especially in industrial settings with programmable logic controllers (PLCs). Advanced Message Queuing Protocol (AMQP) was developed for secure data transmission in financial services, offering robust security features including advanced authentication. Process Field Network (PROFINET) is designed for industrial automation, facilitating real-time data exchange between devices over Ethernet. Finally, Controller Area Network (CAN) bus, developed by Bosch, is widely used in vehicles and industrial systems to enable communication between sensors and devices without the need for an intermediary device. Each protocol is tailored for specific applications, ensuring efficient and secure communication in diverse IoT and industrial environments.

The transport layer faces many security challenges as it acts as middleware between the network and IoT systems. Therefore, it is crucial to implement security techniques to ensure confidentiality, authentication, and integrity.

\subsection{Semantic reasonings layer} 
\begin{figure}[t]
\centering
\includegraphics[width=0.25\textwidth]{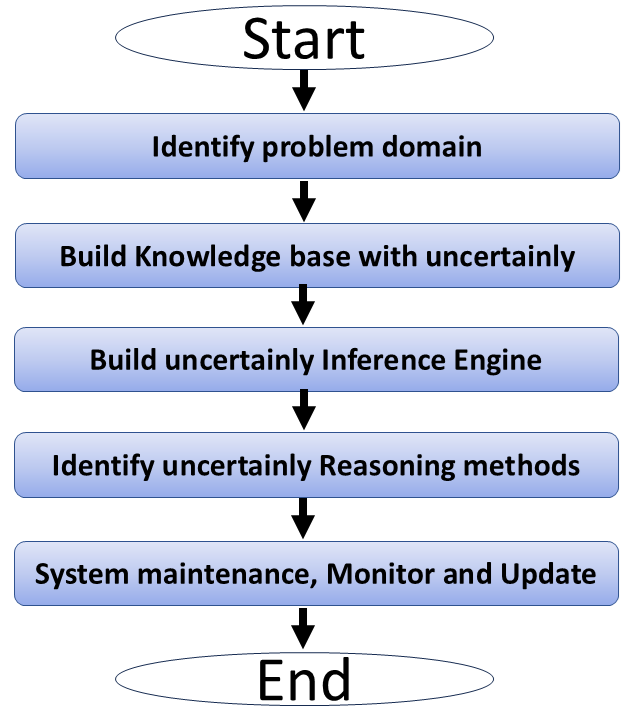}
\caption{Automatic reasoning system with uncertainly}\label{fig:uncertainity}
\end{figure}

\vspace{-4pt}
After the collected data are transmitted to other systems, an expert system is used to infer new knowledge from the transmitted data. Since the collected data may be incomplete and lack some information, an automated reasoning expert system with uncertainty techniques is developed in this layer to enhance the system with semantic completeness. Figure \ref{fig:uncertainity} shows the flowchart depicting the general steps to design an expert system capable of inferring new data from incomplete information and data collected from different sources.

Figure \ref{fig:uncertainity} shows five steps: identify the problem domain, build a knowledge base system, build an inference engine, identify uncertainty reasoning methods, and system maintenance, monitoring, and updating to develop an expert system with uncertain concepts. These steps are needed to create an expert system capable of automatic reasoning with uncertainty methods to infer new knowledge from the collected data.
In the identify problem domain step, the problem that the expert system aims to solve and its purpose are defined. Because IoT devices extract raw, incomplete data, the expert system will depend on uncertainty techniques. Therefore, it is crucial to identify the types of uncertainty the expert system may face.

In building knowledge base system step, the rules and facts about the domain must be entered and stored in the knowledge base by an expert. If new data are collected from sensors, they must also be stored as facts about the domain. The process of entering the rules must include methods for encoding uncertainty within the system. For example, experts may use probabilistic rules and fuzzy sets to incorporate uncertainty concepts.
In the build uncertainty inference engine step, an inference engine must be developed to execute the appropriate rules in the appropriate situations, enabling the expert system to respond appropriately. The inference engine must support, handle, and understand uncertainty or incomplete information. This inference engine must also provide explanations and reasons for choosing specific rules to execute. In uncertain cases, if the inference engine chooses to execute specific rules with a certain probability, it must output the probability of choosing this rule and the alternative rules along with their probabilities.

In the identifying uncertainty reasoning methods step, two different types of reasoning methods are used to infer new knowledge from existing data: traditional reasoning methods and uncertainty reasoning methods. Two types of traditional reasoning methods, rule-based reasoning and case-based reasoning, are used in the suggested framework to extract new knowledge without incorporating uncertainty concepts. Rule-based reasoning uses if-then rules to infer new knowledge, while case-based reasoning relies on past experiences and cases to infer new knowledge. In the context of uncertainty, this paper suggests using three types of reasoning methods: fuzzy logic, Dempster-Shafer theory, and Bayesian networks.
In the system maintenance, monitoring, and updating step, the performance of the expert system must be monitored, and the knowledge base updated when new knowledge becomes available. In this step, the expert system is also evaluated to show how effectively it handles uncertainty. Ongoing maintenance is required to keep the knowledge base current as new information becomes available.
\subsection{Application layer} \vspace{-4pt}
After the expert system extract new knowledge, the data and new knowledge are sent to the application layer. In the application layer, Graphical User Interfaces (GUIs) are implemented and designed to help individuals communicate with IoT devices. Through the GUI, users can monitor their systems and view the environmental measurements collected by IoT devices. Additionally, the results of the expert system and the probability of choosing the appropriate rule are also displayed to users through the GUI.

\section{Agriculture Case study}
\label{sec:Agriculture} \vspace{-4pt}

\begin{figure*}[t]
\centering
\includegraphics[width=0.95\textwidth]{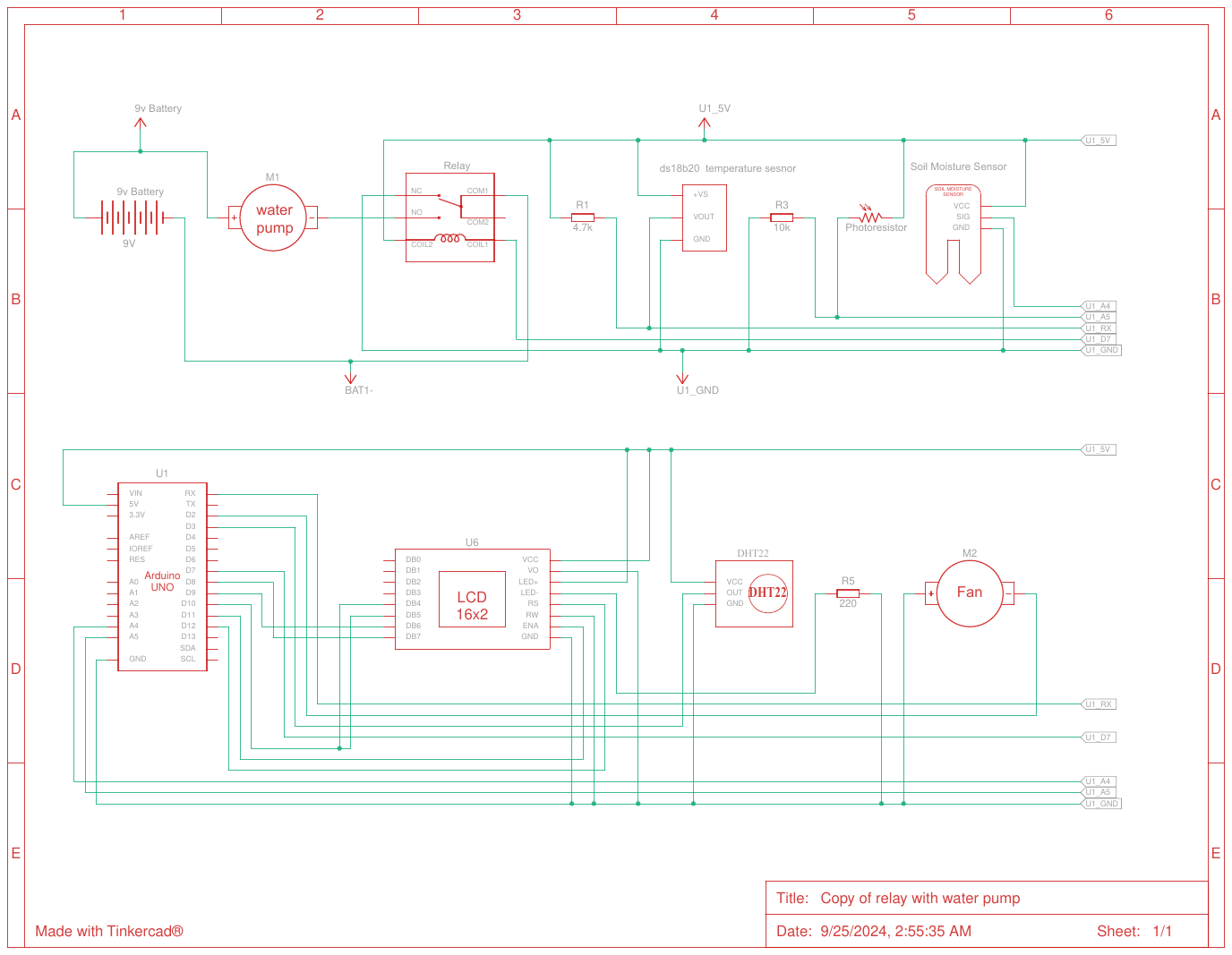}
\caption{Proposed Systematic diagram}\label{fig:cad}
\end{figure*}

Figure \ref{fig:cad} shows a circuit diagram that represents an automated irrigation and climate control system using an Arduino Uno. It is powered by a 9V battery, which supplies energy to a water pump controlled via a relay. The system includes various sensors: a soil moisture sensor for monitoring soil hydration, a DS18B20 and a DHT22 sensor for measuring temperature and humidity, and a photoresistor for detecting light intensity. Soil pH sensor is optional. These sensors are connected to the Arduino, which processes the data and controls actuators like the water pump and a fan based on the readings. An LCD 16x2 display is also included to display real-time information from the sensors. The setup is designed to automate environmental monitoring and watering tasks.

\begin{figure}[t]
\centering
\includegraphics[width=0.5\textwidth]{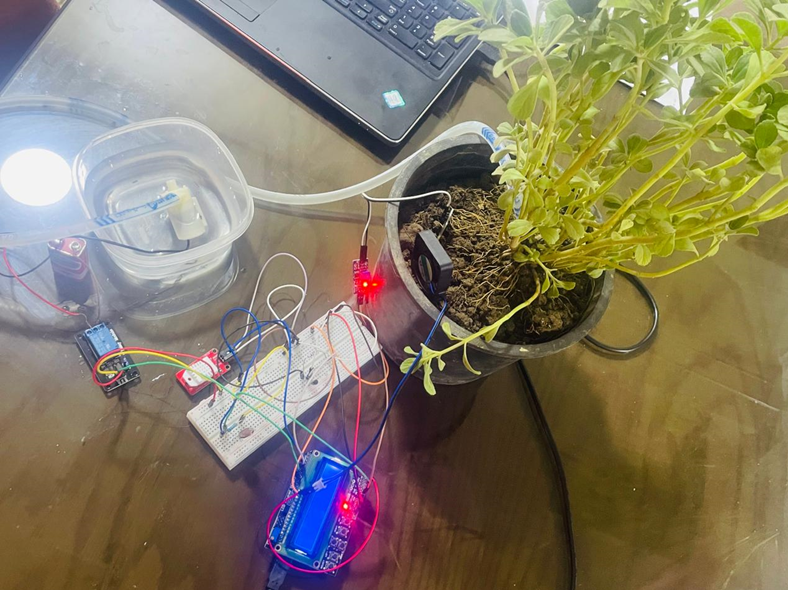}
\caption{Realization of proposed system}\label{fig:real}
\end{figure}

Figure \ref{fig:real} shows the realization of the proposed system. It involves a plant setup with some electronic components for monitoring and automating plant care, sensors and an Arduino microcontroller.

Table \ref{tbl:annotation} provides semantic annotation using interoperable values.  The ambient temperature, measured by a temperature sensor, is recorded at 36.78°C, indicating the surrounding environmental temperature. The air humidity is 68.49\%, which reflects the moisture content in the air. A soil moisture sensor reports a value of 23.45\%, providing insights into the water content present in the soil. The ambient light intensity is measured at 281.40 Lux, revealing the level of illumination in the area. Finally, the soil acidity level, indicated by the soil pH sensor, stands at 5.90, which reflects the current pH balance of the soil. These sensors collectively provide key environmental data for monitoring and analysis.
Based on the sensor readings, several actions can be taken to optimize the environment as shown in Table \ref{tbl:actions}. Given the soil moisture content of 23.45\%, irrigation should be initiated to maintain optimal hydration for plant growth. With the temperature at 36.78°C, the cooling system should be activated to prevent heat stress. Since the light intensity is measured at 281.40 Lux, turning on the grow lights might be necessary to ensure adequate lighting for photosynthesis. Additionally, with the soil pH reading of 5.90, additives may be applied to adjust the pH level, ensuring a balanced soil condition for healthy plant development.

%Table 2 Semantic Annotation
%Sensor	Interoperable Value	Unit	Meaning
%temperature	36.78	Celsius	ambient temperature
%humidity	68.49	%	air humidity
%soil_moisture	23.45	%	soil moisture content
%light_level	281.40	Lux	ambient light intensity
%soil_ph	5.90	pH	soil acidity level

%\begin{table}[ht]
%\centering
%\begin{tabular}{|>{\raggedright}p{4cm}|>{\raggedright}p{4cm}|>{\raggedright}p{2cm}|>{\raggedright}p{6cm}|}
%\hline
%\textbf{Sensor} & \textbf{Interoperable Value} & \textbf{Unit} & \textbf{Meaning} \\
%\hline
%temperature & 36.78 & Celsius & ambient temperature \\
%\hline
%humidity & 68.49 & \% & air humidity \\
%\hline
%soil\_moisture & 23.45 & \% & soil moisture content \\
%\hline
%light\_level & 281.40 & Lux & ambient light intensity \\
%\hline
%soil\_ph & 5.90 & pH & soil acidity level \\
%\hline
%\end{tabular}
%\caption{Semantic Annotation}
%\end{table}

\begin{table*}[h]
    \centering
    \caption{Semantic Annotation} \label{tbl:annotation}
    \begin{tabular}{@{}lccc@{}}
        \toprule
        Sensor          & Interoperable Value & Unit     & Meaning                   \\ \midrule
        Temperature      & 36.78               & Celsius  & Ambient temperature        \\
        Humidity         & 68.49               & \%       & Air humidity              \\
        Soil Moisture    & 23.45               & \%       & Soil moisture content      \\
        Light Level      & 281.40              & Lux      & Ambient light intensity    \\
        Soil pH         & 5.90                & pH       & Soil acidity level         \\ \bottomrule
    \end{tabular}
\end{table*}

%%Table 3 Application Actions
%%Action
%%Irrigate the field
%%Activate cooling system
%%Turn on grow lights
%%Adjust soil pH with additives

%\begin{table}[ht]
%\centering
%\begin{tabular}{|>{\raggedright}p{8cm}|}
%\hline
%\textbf{Action} \\
%\hline
%Irrigate the field \\
%\hline
%Activate cooling system \\
%\hline
%Turn on grow lights \\
%\hline
%Adjust soil pH with additives \\
%\hline
%\end{tabular}
%\caption{Application Actions}
%\end{table}

\begin{table}[h]
    \centering
    \caption{Actions to Take} \label{tbl:actions}
    \begin{tabular}{@{}l@{}}
        \toprule
        Action                                \\ \midrule
        Irrigate the field                    \\
        Activate cooling system               \\
        Turn on grow lights                   \\
        Adjust soil pH with additives         \\ \bottomrule
    \end{tabular}
\end{table}

Table \ref{tbl:recs} provides a semantic reasoning example. The current environmental conditions trigger specific actions for optimization. Since the soil moisture is below 30\% at 23.45\%, it is recommended to irrigate the field to maintain adequate soil moisture levels. Additionally, the temperature exceeds 35°C, with a recorded value of 36.78°C, making it necessary to activate the cooling system to regulate the temperature. The light level is below the threshold of 300 Lux, measured at 281.40 Lux, suggesting the need to turn on the grow lights to ensure sufficient illumination for plant growth. Lastly, the soil pH of 5.90 is outside the optimal range of 6.0 to 7.5, which indicates the soil acidity level needs adjustment using pH-balancing additives.

%Table 4 Semantic Reasoning
%Condition	Explanation	Recommendation
%Soil moisture < 30% 	Soil moisture is 23.45% 	Irrigate the field
%Temperature > 35°C	Temperature is 36.78°C.	Activate cooling system
%Light level < 300 Lux	Light level is 281.40 Lux.	Turn on grow lights
%Soil pH out of range (6.0-7.5)	Soil pH is 5.90.	Adjust soil pH with additives
%
%\begin{table}[ht]
%\centering
%\begin{tabular}{|>{\raggedright}p{4cm}|>{\raggedright}p{6cm}|>{\raggedright}p{6cm}|}
%\hline
%\textbf{Condition} & \textbf{Explanation} & \textbf{Recommendation} \\
%\hline
%Soil moisture < 30\% & Soil moisture is 23.45\% & Irrigate the field \\
%\hline
%Temperature > 35°C & Temperature is 36.78°C & Activate cooling system \\
%\hline
%Light level < 300 Lux & Light level is 281.40 Lux & Turn on grow lights \\
%\hline
%Soil pH out of range (6.0-7.5) & Soil pH is 5.90 & Adjust soil pH with additives \\
%\hline
%\end{tabular}
%\caption{Semantic Reasoning}
%\end{table}
\begin{table*}[h]
    \centering
    \caption{Semantic Reasoning Example} \label{tbl:recs}
    \begin{tabular}{@{}lll@{}}
        \toprule
        Condition                                & Explanation                          & Recommendation                      \\ \midrule
        Soil moisture less than 30\%                    & Soil moisture is 23.45\%           & Irrigate the field                  \\
        Temperature greater than 35°C                       & Temperature is 36.78°C              & Activate cooling system             \\
        Light level less than 300 Lux                   & Light level is 281.40 Lux           & Turn on grow lights                 \\
        Soil pH out of range (6.0-7.5)         & Soil pH is 5.90                    & Adjust soil pH with additives       \\ \bottomrule
    \end{tabular}
\end{table*}

Fuzzy logic uses membership functions that consider each event by some degree of occurrence rather than binary true or false values, which helps the expert system handle imprecise concepts.
Fuzzy membership functions describe how each point in the input space (i.e., the agriculture domain) is mapped to a degree of membership between 0 and 1. For the soil moisture example, the fuzzy sets are described by triangular membership functions (trimf), which are defined by three parameters: the left, center, and right points. Here are the equations for each fuzzy membership function:
Low Soil Moisture (Triangular Membership Function), defined by the range [0, 0, 30]:
\begin{equation}
    \mu_{low}(x) =
    \begin{cases}
    1 & \text{if } x \leq 0 \\
    \frac{30 - x}{30} & \text{if } 0 < x < 30 \\
    0 & \text{if } x \geq 30
    \end{cases}
\end{equation}

Adequate Soil Moisture (Triangular Membership Function), defined by the range [20, 50, 80]:
\begin{equation}
\mu_{\text{adequate}}(x) =
\begin{cases}
1 & \text{if } x \leq 20 \\
\frac{30 - x}{30} & \text{if } 20 < x < 50 \\
0 & \text{if } x \geq 50
\end{cases}
\end{equation}

\begin{figure}[t]
\centering
\includegraphics[width=0.5\textwidth]{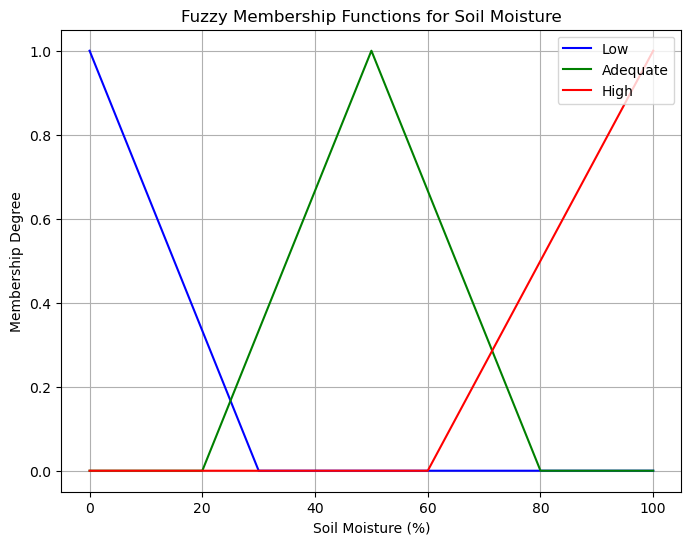}
\caption{Fuzzy Membership Functions for Soil Moisture}\label{fig:fuzzy}
\end{figure}

High Soil Moisture (Triangular Membership Function), defined by the range [60, 100, 100]:
\begin{equation}
\mu_{\text{high}}(x) =
\begin{cases}
0 & \text{if } x \leq 60 \\
\frac{x - 60}{40} & \text{if } 60 < x < 100 \\
1 & \text{if } x = 100
\end{cases}
\end{equation}
Figure \ref{fig:fuzzy} shows fuzzy membership functions for soil moisture which are defined across three distinct categories. For low soil moisture, the membership degree is 1 when the moisture level is at 0\%, indicating complete membership in this category. As soil moisture increases and reaches 30\%, the membership degree linearly decreases to 0, remaining at this level for any moisture content beyond 30\%. In the case of adequate soil moisture, the membership degree starts at 0\% when the moisture level is at 20\% and linearly increases to 1 at 50\%, indicating full membership. However, this degree decreases back to 0 by 80\%. Lastly, for high soil moisture, the membership degree remains at 0 below 60\%, indicating no membership in this category, but increases to 1 as the moisture level reaches 100\%, signifying full membership for high moisture content.

Dempster-Shafer theory is important for IoT applications because it aims to develop new knowledge from data that come from different sources.
In the context of Dempster-Shafer theory applied to soil moisture, we can define the key components and equations that represent how to combine beliefs from different sources regarding the state of soil moisture (e.g., "Low," "Adequate," and "High").
A basic belief assignment (BBA) is a function that assigns a degree of belief to each proposition based on the evidence provided. For soil moisture, the BBA can be represented as follows:
\begin{equation}
m: 2^\theta \to [0, 1]
\end{equation}

where $\theta$=\{Low, Adequate, High\}

To combine the beliefs from two sources A and B, we use Dempster's Rule of Combination. The combined belief $m_C$ for a proposition can be calculated as follows:
\begin{equation}
m_C(A) = \frac{1}{1-K} \sum_{B \cap C = A} m_A(B) \cdot m_B(C)
\end{equation}
where K is the conflict measure, that accounts for the degree of conflict between the two sources. It quantifies how much evidence is incompatible or contradictory. If there is no conflict, K will be close to 0; if there is significant conflict,  K will be higher. K is defined as:
\begin{equation}
K = \sum_{A \cap B = \emptyset} m_A(A) \cdot m_B(B)
\end{equation}

\begin{figure}[t]
\centering
\includegraphics[width=0.5\textwidth]{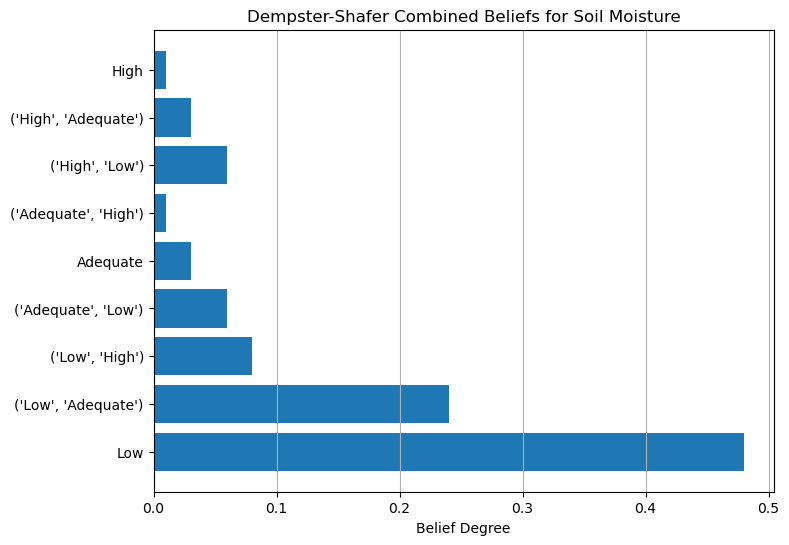}
\caption{Dempster-Shafer Combined Beliefs for Soil Moisture}\label{fig:Dempster}
\end{figure}

After calculating the combined beliefs using Dempster's rule, it’s important to normalize the results to ensure that the total belief sums to 1:
\begin{equation}
m_C(A) = \frac{m_C(A)}{1-K}
\end{equation}
Using Dempster's rule, the combined beliefs for each proposition would be calculated, considering the conflict K and normalizing the results. The specific values would depend on the actual calculations performed using the equations above. Figure \ref{fig:Dempster} shows how Dempster-Shafer theory combines beliefs by providing a structured way to combine uncertain evidence from multiple sources, allowing for reasoning about the soil moisture levels based on varying degrees of belief. The use of BBAs, combination rules, and conflict measures facilitates a nuanced understanding of the information gathered from different observations.

\begin{figure}[t]
\centering
\includegraphics[width=0.5\textwidth]{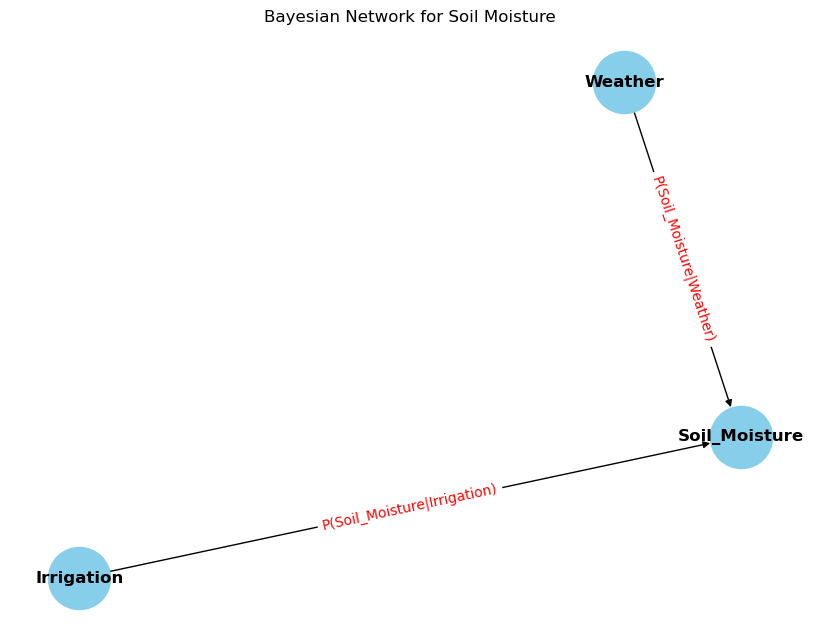}
\caption{Bayesian Network for Soil Moisture}\label{fig:bayes}
\end{figure}

Bayesian networks, based on Bayes' rule, estimate the probability of certain outcomes based on related events. Figure \ref{fig:bayes} shows the Bayesian network for soil moisture management which consists of three nodes: Weather, Irrigation, and Soil Moisture. The Weather node indicates the likelihood of rain, with probabilities set at 70\% for rain and 30\% for no rain. The Irrigation node reflects the irrigation status, with a 60\% chance of being on and a 40\% chance of being off. The Soil Moisture node assesses the soil's moisture level based on the conditions from the other two nodes, incorporating conditional probabilities for low, adequate, and high moisture levels depending on the weather and irrigation states. For instance, when it is raining and irrigation is active, there is a high probability of adequate moisture, while low moisture is more likely when it is not raining and irrigation is off. This structured approach enables informed decision-making in agricultural practices, optimizing water usage based on environmental conditions. The network can be visualized using directed graphs, providing clear insights into the relationships and dependencies among the variables involved in soil moisture management.

\begin{figure}[t]
\centering
\includegraphics[width=0.25\textwidth]{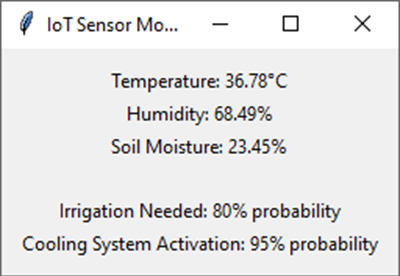}
\caption{Graphical User Interfaces Alert}\label{fig:gui}
\end{figure}

Based on the probabilistic reasoning presented above, the system is able to suggest actions such as the actions presented in Figure \ref{fig:gui}.

\section{Conclusion and Future Work}
\label{sec:Conclusion} \vspace{-4pt}
The Internet of Things (IoT) has a significant impact and has advanced many applications, such as smart cities, smart homes, agriculture, and healthcare systems around the world. Despite the substantial impact of IoT, it still suffers from incomplete data collection and the inability of IoT devices and sensors to fully understand the data. Therefore, this paper proposes a new real-time framework that includes three different semantic layers to enable IoT devices and sensors to understand the meaning and source of the collected data. This framework can be used in any IoT application as it relies on techniques suitable for environments with dynamic behavior. The framework consists of six layers: perception, semantic annotation and preprocessing, semantic interoperability and ontologies, transportation, semantic reasoning, and application.

In the perception layer, sensors collect data from the environment in the form of voltage. These sensors send the collected data to a microprocessor or microcontroller device capable of processing the data and converting it to meaningful information in the semantic annotation and preprocessing layer. In this layer, metadata is added to the collected raw data, such as the purpose of the sensors, the sensor ID number, and the application the sensors are used for. Descriptions of the collected data are also added to the raw data. Two semantic algorithms are proposed in the semantic interoperability and ontologies layer: the interoperability semantic algorithm, which facilitates data exchange by standardizing file types to allow data to be sent in any format, and the synonym identification algorithm, which finds synonyms of keywords to help individuals understand unfamiliar concepts. In the transportation layer, the raw data and metadata are sent to other IoT devices or cloud computing platforms using various techniques such as WiFi, Zigbee networks, Bluetooth, and mobile communication networks. Because the IoT collected data is often incomplete, a semantic reasoning layer is suggested to infer new knowledge from the existing data. In the semantic reasoning layer, a new expert system based on uncertainty concepts is proposed to build a knowledge base system using fuzzy logic, Dempster-Shafer theory, and Bayesian networks to help derive new knowledge in both certain and uncertain cases.

Finally, a Graphical User Interface (GUI) is proposed in the application layer to help users communicate with and monitor IoT sensors, IoT devices, and the new knowledge inferred, including the probability of the inferred knowledge and other alternative knowledge with their probabilities, as well as metadata of each sensor. This framework provides a robust solution for managing IoT data, ensuring semantic completeness, and enabling real-time knowledge inference. The integration of uncertainty reasoning methods and semantic interoperability techniques makes this framework a valuable tool for advancing IoT applications in various domains.

One possible future direction is to extend the proposed framework, providing intricacies of implementing each layer.
%For example, ontologies can be built using Prot\'eg\'e which is an open-source ontology editor and a knowledge management system.
Another possible direction is securing the IoT system. The most possible security approach is to combine blockchain \cite{el2019spainchain}, or using cyber-physical systems \cite{moskvins2022intelligent}.
Another future work may extend the scale of the proposed system using wireless sensor network \cite{dharani2020study}.

\end{document}